\documentclass{article}
\usepackage{spconf}
\usepackage{graphicx}
\usepackage{amsmath}
\usepackage[pagebackref,breaklinks,colorlinks]{hyperref}
\usepackage{amssymb}
\usepackage{booktabs}
\usepackage[capitalize]{cleveref}
\usepackage{orcidlink}
\usepackage{lipsum} 
\usepackage{mathtools}
\usepackage{multirow}
\usepackage{color, colortbl}
\usepackage{diagbox}
\usepackage{soul}
\usepackage{xcolor}
\usepackage{enumitem}
\usepackage{tabularx} 
\usepackage{svg}
\usepackage{adjustbox}
\usepackage{lineno}

\usepackage{enumitem}
\setlist{nosep, leftmargin=14pt}
\newlength{\bibitemsep}
\newlength{\bibparskip}
\let\oldthebibliography\thebibliography
\renewcommand\thebibliography[1]{%
  \oldthebibliography{#1}%
  \setlength{\parskip}{\bibitemsep}%
  \setlength{\itemsep}{\bibparskip}%
}

\usepackage{mwe} % to get dummy images

\title{C-arm Guidance: A SELF-SUPERVISED APPROACH TO AUTOMATED POSITIONING DURING STROKE THROMBECTOMY}
 \name{A. Arrabi$^{1\star}$, J. Jung$^{1\star}$, J. Le$^2$, A. Nguyen$^2$, J. Reed$^2$, E. Stahl$^2$, N. Franssen$^2$, S. Raymond$^3$, S. Wshah$^1$}
 
 \address{$^1$ University of Vermont, Department of Computer Science, Burlington VT, USA \\
     $^2$Univeristy of Vermont Medical Center, Burlington VT, USA \\
     $^3$ Cleveland Clinic, Neurological Institute, Cleveland OH, USA \\
     \small $^\star$ Equal contribution
     \vspace{-0.25cm}
     }

\begin{document}
\maketitle
\begin{abstract}
Thrombectomy is one of the most effective treatments for ischemic stroke, but it is resource and personnel-intensive. We propose employing deep learning to automate critical aspects of thrombectomy, thereby enhancing efficiency and safety. In this work, we introduce a self-supervised framework that classifies various skeletal landmarks using a regression-based pretext task. Our experiments demonstrate that our model outperforms existing methods in both regression and classification tasks. Notably, our results indicate that the positional pretext task significantly enhances downstream classification performance. Future work will focus on extending this framework toward fully autonomous C-arm control, aiming to optimize trajectories from the pelvis to the head during stroke thrombectomy procedures. All code used is available at \url{https://github.com/AhmadArrabi/C_arm_guidance}
\end{abstract}
\vspace{-8pt}
\section{Introduction}
\label{sec:intro}
\vspace{-0.25cm}
%Procedural fluoroscopy uses X-ray imaging to guide therapeutic interventions and is used across a wide range of surgical and interventional specialties from cardiology to endovascular neurosurgery.
Many of the most time-sensitive operations for life-threatening conditions such as stroke and trauma require rapid and accurate positioning of fluoroscopy equipment. These procedures are increasingly performed by less experienced operators in low-resource settings~\cite{stein2021correlations}, creating a need for technologies to improve efficiency and safety.

During stroke treatment, the operator first gains vascular access at the groin and then drives the biplane, two orthogonal X-ray detectors, from a position over the pelvis up to the brain. Fluoroscopy during positioning exposes the patient and operator to extra radiation and takes precious minutes and attention away from other critical tasks. Semi- or automated positioning could reduce the time and radiation required~\cite{desilva2018}, and free the operator to focus on other critical tasks such as device preparation. Thus, we propose the use of deep learning algorithms to assist operators with fluoroscopy operation, particularly in solo- or low-volume settings.

Automated X-ray positioning has been proposed for orthopedic applications~\cite{martin_vicario_automatic_2022,kausch_c-arm_2022}, focusing primarily on specific body parts, e.g., lumbar spine, proximal femur~\cite{kausch_toward_2020}, knee~\cite{kausch_2023}, and pelvis~\cite{hooman_2020}. However, %it has not been employed for stroke because of challenges 
its application in stroke treatment remains unexplored due to several challenges: the larger range of motion from the groin to the head, the precise positioning required for cerebral angiographic views, and the broad array of landmarks and radio-dense distractors. 
%To the best of our knowledge, this is the first study that tackles fluoroscopy positioning in stroke treatment, incorporating a larger anatomical region, continuously capturing X-ray images from the pelvis to the head.}

We tackle fluoroscopy positioning in stroke thrombectomy, incorporating a larger anatomical region from the pelvis to the head. Our approach tackles two key tasks, 1) a \textbf{classification task} that aids in accurate interpretation of the current biplane position relative to the patient %based on a minimum number of images, 
i.e. \textit{what body part are we looking at?}, and 2) a \textbf{regression task} that estimates the biplane position based on the X-ray input, i.e., \textit{where are we located?} Our framework presents a step towards broader automation in fluoroscopy control, to ultimately predict the needed trajectory from the current location to the target, i.e. \textit{how do I get to the head from a given location?}
%from any body parts, covering from the femoral area up to the head.

%prediction of the needed trajectory to go from the current location to the intended location, i.e. how do I get to the head from the pelvis?

This work presents a self-supervised learning framework to classify 20 skeletal landmarks between the pelvis and the head. Our approach is centered around a positional understanding pretext task, where the model estimates the location of a given X-ray image. Additionally, we embed patient demographic data, that may reveal variations in skeletal structure across patients, into the learned features. Leveraging DeepDRR~\cite{unberath2018deepdrrcatalystmachine}, we developed a custom graphical user interface (GUI) to annotate and generate synthetic X-ray images from computed tomography (CT) scans. We collected two datasets: one includes evenly distributed X-ray images from CTs, for regression, and another, consists of ground truth annotations for classification. All CTs were downloaded from the New Mexico Decedent Image Database (NMDID)~\cite{edgar2020nmdid}.
%collected two datasets: one  X-ray images generated by DeepDRR~\cite{unberath2018deepdrrcatalystmachine} for a self-supervised regression task, and another consisting of ground truth annotations from a custom GUI that maps CT coordinates to X-ray coordinates for a downstream classification task. The CT scans were sourced from the New Mexico Decedent Image Database (NMDID)~\cite{edgar2020nmdid}.}

%Leveraging DeepDRR~\cite{unberath2018deepdrrcatalystmachine}, we developed a custom graphical user interface (GUI) to generate synthetic X-ray images from computed tomography (CT) scans, enabling the collection of two datasets. The first dataset includes over 145,338 unannotated X-ray images from CT volumes, used for the self-supervised pretext task (regression). The second dataset, 8080 annotated images by medical residents, serves as the ground truth for the downstream task (classification). All CT scans were downloaded from the New Mexico Decedent Image Database (NMDID)~\cite{edgar2020nmdid}.}

Our contributions can be summarized as follows:
\begin{itemize}
    \item We work toward automating fluoroscopy positioning (with broader anatomical coverage) for stroke thrombectomy.
    \item We introduce a self-supervised framework that embeds patient demographic data into its learned features. We find that by leveraging a positional understanding pretext task trained on a large dataset, the performance of the downstream classifier significantly improves. 
    \item We put forward a custom public GUI to simulate C-arm operation, which can be used as an annotation tool.
    % \item We put forward a custom GUI to simulate C-arm operation and generate synthetic X-ray images, which can be used as an annotation tool. \textit{Guidelines of generating our datasets will be made publicly available.}
\end{itemize}

\vspace{-0.25cm}
\section{Dataset}
\label{sec:dataset}
\vspace{-0.15cm}
To generate the Digitally Reconstructed Radiographs (DRRs), we used whole-body CT scans from the New Mexico Decedent Image Database (NMDID)~\cite{edgar2020nmdid}. This type of scan covers all regions of interest in stroke thrombectomy, thereby mirroring the continuous movement of the C-arm in procedures. The DRRs were generated by DeepDRR~\cite{unberath2018deepdrrcatalystmachine}, providing synthetic X-ray images.
%Data for this study was collected in two steps: (1) anatomical positioning data was obtained from CT scans, and (2) Digitally Reconstructed Radiographs (DRRs) were generated, specifically as predicted X-ray images using the deep learning algorithm DeepDRR~\cite{unberath2018deepdrrcatalystmachine}.
  %Additionally, one of the study's objectives is to automatically navigate the mobile C-arm from any arbitrary position to the cephalic region using a Deep Learning algorithm. 
%Additionally, the larger field of view enables better simulation of real-life operations and captures anatomical positioning dependencies across the body.
%The NMDID was initially collected by the Office of the Medical Investigator as part of a 2010 grant (NIJ 2010-DN-BX-K205), aimed at supplementing or replacing autopsies with CT scans in cases involving blunt force trauma, firearm injuries, drug poisoning, and childhood trauma. Individual cases can be requested and downloaded from the NMDID website, with each containing multiple parts and varying specifications of CT scans. While the database provides valuable metadata and whole-body CT scan on decedents, such as height and weight, the images are stored as flat files (DICOM), which poses challenges related to storage and visualization. 

The NMDID is a public database containing full-body CT scans and demographics of over 30,000 decedents. %The raw downloaded data were stored as flat files (DICOM), which poses challenges related to storage and visualization. Thus, specific criteria were applied when selecting CT scans. As mentioned, each case contains images with varying specifications. 
We selected CT scans with sharper filters (labeled “BONE\_”) for faster X-ray synthesis with DeepDRR. To ensure full-body coverage, we chose two scan sets, one with arms crossed (labeled “\_H-N-UXT\_3X3”), and one with arms raised (labeled “\_TORSO\_3\_X\_3”). %In both cases, “3x3” indicates 3 mm slices with 3 mm spacing and no overlap. After downloading the scans flat files (DICOM), %we converted them to NIFTI (.nii) format and we compressed them into .nii.gz files to reduce storage size.

\begin{figure}[!b]

\centering
    \centering
    \fbox{\includegraphics[width=0.95\linewidth, trim={0cm 0cm 0cm 0cm}, clip]{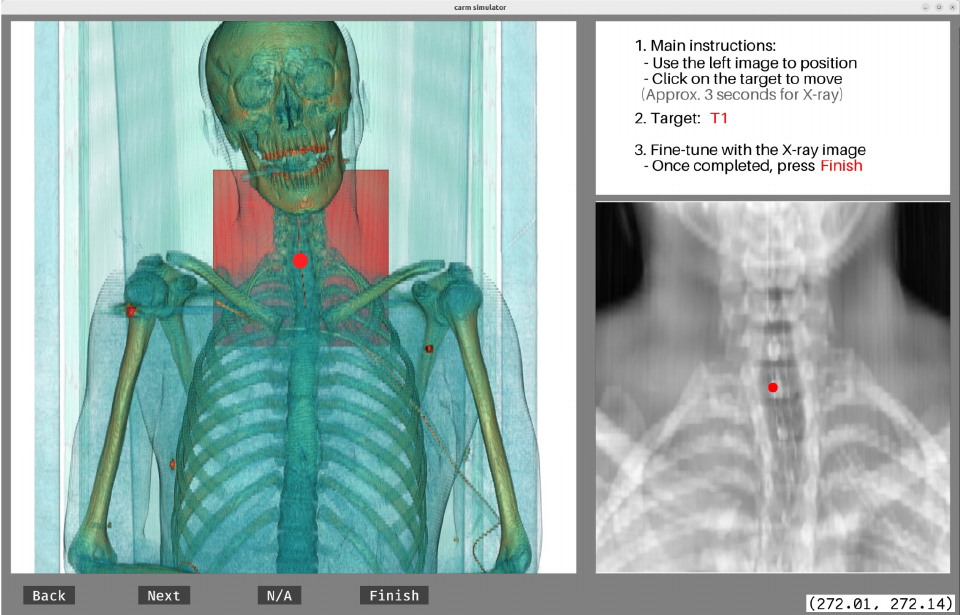}}
    \caption{GUI visualization. The annotators were given direct instructions on collecting the 20 landmarks (top right).}
    \label{fig:gui}
    \vspace{-0.25cm}
\end{figure}

\noindent\textbf{Graphical User Interface (GUI) Annotations}
As shown in \cref{fig:gui}, we developed a custom GUI in Python to facilitate landmark annotation on CT scans. The 3D visualization was implemented using Vedo~\cite{musy2019vedo}, while the GUI was built with Tkinter~\cite{lundh1999tkinter}. The GUI simulates a C-arm, allowing users to annotate specific landmarks by displaying both 3D views of the CT scan and corresponding DRR images. 

%This comprehensive field of view ensures accurate landmark annotation. Users can also control both the CT scan and X-ray simulation at the millimeter level, capturing the precise coordinates of each landmark. These recorded coordinates are then stored in a database for future reference. 
%\vspace{-1cm}
\begin{figure}[!t]
    \centering
    \begin{minipage}[b]{0.7\linewidth}
        \centering
        \centerline{\fbox{\includegraphics[width=0.98\linewidth]{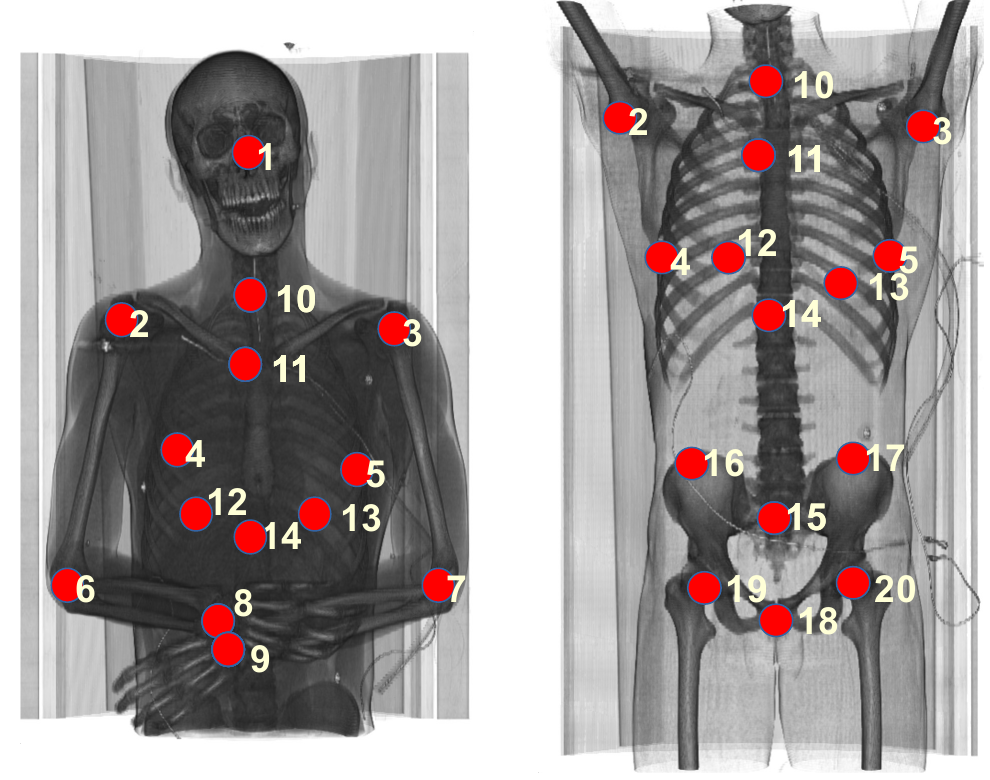}}}
    \end{minipage}
    \begin{minipage}[b]{0.28\linewidth}
        \centering
        \centerline{\includegraphics[width=0.98\linewidth]{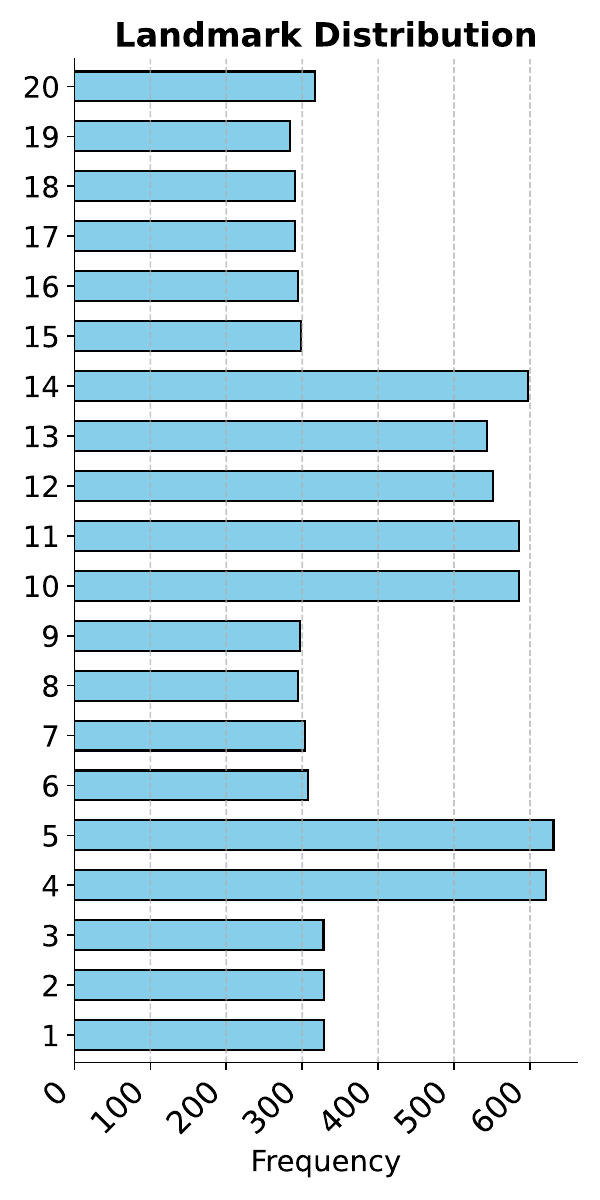}}
    \end{minipage}
    \caption{Landmarks of spatial locations across the skeleton (left), and their distribution in our annotated dataset (right).}
    \vspace{-0.5cm}
\label{fig:landmarks_dist}
\end{figure}

%To visualize CT scans for landmark annotation, a customized graphical user interface (GUI) was developed using Python. The 3D object visualization was implemented with the Vedo module, while the GUI itself was built with Tkinter. The primary function of the GUI is to simulate a C-arm, allowing users to annotate specific landmarks. It is designed to display both 3D views of the CT scan and the corresponding X-ray image generated by DeepDRR, providing a comprehensive field of view for accurate annotation. Additionally, the GUI collects the coordinate positions of each landmark in every available CT scan and stores this data in a database.
%\textbf{"If I have space, I would include descriptions for each landmark."}
\noindent\textbf{Annotated Classification Dataset}
Utilizing the GUI, we collected annotations of 270 cases (decedents), 47 of which were annotated by radiology residents and the rest were from a non-physician. In total, 8,080 annotations of 20 landmarks were made, \cref{fig:landmarks_dist} illustrates the location of each landmark and their distribution. Some landmarks are more frequent than others as they exist in both CT scans.

Operators were instructed to click on the position of the assigned landmark. The GUI recorded the C-arm's coordinates through translation only, i.e., 2 degrees of freedom (no rotation, with the z-coordinate fixed). After positioning the C-arm over the landmark, the GUI generated the DRR, which operators could manually fine-tune its position. The following anatomic landmarks were labeled: 1) skull, 2–3) humeral heads, 4–5) scapulas, 6–7) elbows, 8–9) wrists, 10) T1, 11) carina, 12–13) hemidiaphragms, 14) T12, 15) L5, 16–17) iliac crests, 18) pubic symphysis, and 19–20) femoral heads.
%Utilizing the GUI, seven operators contributed to the annotation process, including six radiology residents and one non-resident. 270 cases (decedents) were annotated, with 3,798 annotations from 256 lower CT scans and 4,282 annotations from 262 upper CT scans. In total, 8,080 annotations of 20 landmarks were made, with each landmark having a varying number of annotations. \cref{fig:landmarks_dist} illustrates the location of each landmark and their corresponding distribution in the annotated dataset. %The dataset consists of 1,181 annotations collected by residents and 6,899 by the non-resident.
%Operators were instructed to click on the position of the assigned landmark. The GUI recorded the C-arm's coordinates through translation only, i.e., 2 degrees of freedom (no rotation, with the z-coordinate fixed based on the 3D objects). After positioning the C-arm over the landmark, the GUI generated the DRR, which operators could manually fine-tune its position. %Once operators confirmed the position, they proceeded to the next landmark by clicking the 'Next' button. 
%The following anatomic landmarks were labeled: 1) skull, 2–3) humeral heads, 4–5) scapulas, 6–7) elbows, 8–9) wrists, 10) T1, 11) carina, 12–13) hemidiaphragms, 14) T12, 15) L5, 16–17) iliac crests, 18) pubic symphysis, and 19–20) femoral heads.
%they would manually capture the X-ray image with a click.
%\textbf{"If I have space, I would include distributions for each landmark."}

\noindent\textbf{Regression Dataset}
We utilized 392 cases, of which 270 overlap with the ones described above, to generate a larger unannotated dataset for the regression pretext task. We densely sampled CT scans by defining a uniform grid over them, with points spaced 30 mm apart in both vertical and horizontal directions. X-ray images were sampled at these grid points, ensuring systematic and comprehensive coverage of anatomical features. We hypothesize that this approach enables the model to effectively learn the positional relationships of landmarks within the body. A total of 145,338 images were collected, with 76,578 from the one with arms raised CTs and 68,760 from the arms crossed. \textit{As shown in \cref{sec:experiments}, the classification and regression tasks were trained on the same cases (decedents), ensuring no overlap between the training and testing sets.}

\vspace{-0.25cm}
\section{Methodology}
\label{sec:methodology}
\vspace{-0.25cm}
Acquiring large labeled datasets in medical imaging poses significant challenges, e.g., privacy concerns of protected health information, and the need for domain experts for annotations~\cite{self_medical}. Self-supervised learning addresses these issues by learning meaningful data representations from large datasets without explicit human labels~\cite{self_survey}. A model is first trained on a pretext task to capture the underlying structure of the data and then fine-tuned for the target downstream task, which typically has limited labeled data.

Selecting an appropriate pretext task plays an important role in shaping the learned representations during training. In this work, we propose a regression-based pretext task, where the model predicts the spatial location of a given X-ray image. We hypothesize that this task enables the model to develop a comprehensive understanding of full-body anatomical positioning, particularly of the skeletal structure. Additionally, we propose to embed patient's demographic data into the model, incorporating age, sex, cadaver height, and weight to enrich the model's learned features.

As illustrated in \cref{eq:main_model}, we can formalize our method as follows: Given an input X-ray image $I_{X_{ray}}$, located at position $\textbf{P} = (x,y,z)$ in a predefined coordinate system, and demographic data $s_{stats}$. Our model $g_\theta$ estimates the image location $\textbf{P}$ given both $I_{X_{ray}}$ and $s_{stats}$, where $\theta$ are the network's learnable parameters. \textit{Note that our chosen origin point $O(0,0,0)$ was the top right corner of the CT scan.}
\begin{equation}
    g_\theta(I_{X_{ray}}, s_{stats}) = \hat{\textbf{P}} = (\hat{x}, \hat{y}, \hat{z})
    \label{eq:main_model}
\end{equation}
\begin{figure}[!t]
\vspace{-0.25cm}
\centering
    \centering
    \includegraphics[width=\linewidth]{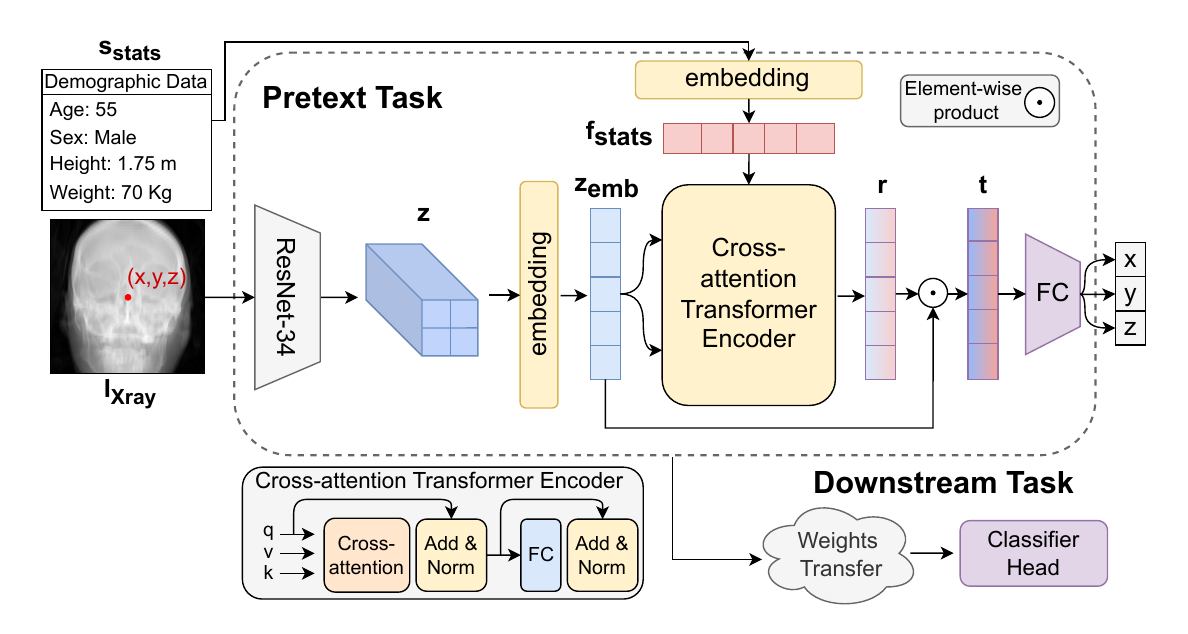}
    \caption{Main architecture of our proposed framework. The learned representations in the regression task are used to fine-tune the downstream classification task.}
    \label{fig:model}
    \vspace{-0.45cm}
\end{figure}
As shown in \cref{fig:model}, $I_{X_{ray}}$ is first projected into a lower-dimensional latent representation $z \in \mathbb{R}^{c\times h\times w}$ using a ResNet-34~\cite{resnet} backbone, where $c$, $h$, and $w$ are the channel, height, and width dimensions, respectively. Then, both $s_{stats}$ and $z$ are projected into continuous vector representations $f_{stats},\ z_{emb} \in \mathbb{R}^D$ through fully connected layers, where $D$ is the embedding dimension. We incorporate cross-attention between these two representations, where the query is derived from $f_{stats}$, and the key and value are obtained from $z_{emb}$. The output of the cross-attention mechanism is further processed by a standard transformer block~\cite{transformer}, producing the vector $r \in \mathbb{R}^{D}$. This vector is expected to have both spatial and demographic information. To further refine the signal and reduce any noise from $f_{stats}$, a skip connection from $z_{emb}$ is added, combining it with $r$ via element-wise multiplication to generate the final refined feature $t \in \mathbb{R}^{D}$. Finally, the regression head consists of a fully connected network with two linear layers that map $t$ to $(\hat{x}, \hat{y}, \hat{z})$. We adopted the Mean Squared Error (MSE) loss to train our model.

After training the regression model, we adapt the architecture for the downstream classification task by replacing the regression head with a classification head. Leveraging the pretrained weights from the pretext task allows the model to preserve rich, domain-specific representations, which enhances its ability to interpret X-ray images. This fine-tuning significantly improves classification performance compared to models initialized with random weights or even pretrained on out-of-domain data, such as ImageNet.

Given that this is a multi-class classification task, we fine-tune the model using the cross-entropy loss. There are several strategies for fine-tuning: one approach involves retraining the entire model, which updates all weights. However, in resource-constrained environments, a more efficient technique known as "linear probing" can be employed. In this method, only the final linear layers are updated and all previous layers are frozen. Linear probing achieves a balance between performance and computational efficiency, as it minimizes the need to retrain the entire model while still adapting to the new task.
\vspace{-8pt}
\vspace{-0.15cm}
\section{Experiments}
\label{sec:experiments}
\vspace{-0.15cm}
\textbf{Implementation Details:} All models were implemented using PyTorch~\cite{pytorch}. We used a batch size of 512 trained across 8 AMD Radeon MI50 GPUs in the regression task. For the classification task, the batch size was 64, trained on a single GPU. We employed the Adam optimizer with a learning rate of 0.0001. All X-ray images were resized to $256 \times 256$ pixels, and random color jittering and posterization were used for data augmentation. The embedding dimension $D$ for the vectors $z_{emb}, r, \text{and } t$  was set to 128.

For evaluation, we benchmark our model on a test set of 54 randomly selected decedents. This split ensures no overlap between the training and testing datasets. For the regression task, we used the MSE between the predicted and the ground truth coordinates, while for classification, we computed micro-averaged precision, recall, and F1-score. Our baseline for both tasks was the modified PoseNet from~\cite{kausch_toward_2020}, a method used in previous automated positioning research.

As shown in \cref{tab:main_result}, our regression model achieved a mean positional error of 4.7 mm, outperforming PoseNet, reducing the error by approximately 1.2 mm. For classification (\cref{tab:classification1}), we retrain the whole network and compared three weight initialization strategies: 1) from the pretext regression task, 2) from ImageNet~\cite{imagenet} pretraining, and 3) random initialization. We also included ViT-base~\cite{vit} and PoseNet as baselines. Our model achieved the best performance across all metrics, with notable improvement when using pretext task weights, highlighting the advantage of domain-specific pretraining over out-of-domain ImageNet weights. Vit-base performed the worst, which we attribute to it requiring more data for optimal performance given it is transformer-based.

Additionally, we explored linear probing by freezing all but the last two linear layers. \cref{tab:classification2} shows a notable performance decline with ImageNet and random weight initializations, demonstrating the benefit of task-specific pretraining.

Finally, we conducted ablation studies to assess the impact of two factors: 1) training fewer layers in linear probing, and 2) removing patient demographics from the model. \cref{tab:ablation} shows that training only one linear layer led to about a 0.2 performance drop, which is expected due to fewer learnable parameters. Furthermore, excluding patient demographic data resulted in a slight performance decline, showing the value of incorporating these features.
\begin{table}[!ht]
    \centering
    \vspace{-0.5cm}
    \small
    \caption{Regression model performance, best results are bold.}
    \begin{tabular}{c|cc}
    \toprule 
    Method & MSE (raw) $\downarrow$ & MSE (mm) $\downarrow$ \\
    \midrule
    ours & \textbf{0.0084} & \textbf{4.7989} \\
    PoseNet~\cite{kausch_toward_2020} & 0.0106 & 6.0717 \\
    \bottomrule
    \end{tabular}
    \label{tab:main_result}
    \vspace{-0.5cm}
\end{table}
\begin{table}[!ht]
    \centering
    %\vspace{-15pt}
    \small
    \caption{The classification results when retraining all layers of our model. $\dag$ indicates initializing the weights from the regression task. $\ddag$ indicates ImageNet initialization, and $\star$ is random weight initialization.}
    \begin{tabular}{c|ccc}
    \toprule 
    Method & Precision $\uparrow$ & Recall $\uparrow$ & F1-score $\uparrow$ \\
    \midrule
    ours$^{\dag}$   & \textbf{0.95} & \textbf{0.95} & \textbf{0.95}  \\
    ours$^{\ddag}$   & 0.94 & 0.93 & 0.93 \\
    ours$^{\star}$  & 0.93 & 0.92 & 0.92  \\
    PoseNet~\cite{kausch_toward_2020}         & 0.91 & 0.91 & 0.91  \\
    ViT-base~\cite{vit}        & 0.79 & 0.77 & 0.77 \\
    \bottomrule
    \end{tabular}
    \vspace{-0.5cm}
    \label{tab:classification1}
\end{table}
\begin{table}[!ht]
    \centering
    %\vspace{-20pt}
    \small
    \caption{The classification results with linear probing on the last two linear layers. A significant gap is present between pretraining on domain-specific and out-of-domain data.}
    \begin{tabular}{c|ccc}
    \toprule 
    Weight Initialization & Precision $\uparrow$ & Recall $\uparrow$& F1-score $\uparrow$ \\
    \midrule
    pretext (regression)       & \textbf{0.83} & \textbf{0.83} & \textbf{0.82}  \\
    ImageNet   & 0.38 & 0.41 & 0.38 \\
    Random  & 0.22 & 0.20 & 0.17  \\
    \bottomrule
    \end{tabular}
    \label{tab:classification2}
    \vspace{-0.5cm}
\end{table}
\begin{table}[!ht]
    \centering
    %\vspace{-15pt}
    \small
    \caption{Ablation studies results. Fine-tuning more linear layers naturally improves performance, and the addition of demographic patient data leads to slight enhancement.}
    \begin{tabular}{c|ccc}
    \toprule 
    Ablation study & Precision $\uparrow$& Recall $\uparrow$& F1-score $\uparrow$\\
    \midrule
    Two layer linear probing & \textbf{0.83} & \textbf{0.83} & \textbf{0.82}  \\
    One layer linear probing   & 0.62 & 0.63 & 0.60 \\
    \midrule
    w/ Patient demographics & \textbf{0.95} & \textbf{0.95} & \textbf{0.95}  \\
    wo/ Patient demographics & \textbf{0.95} & 0.94 & 0.94 \\
    \bottomrule
    \end{tabular}
    \label{tab:ablation}
    \vspace{-0.25cm}
\end{table}
\vspace{-0.6cm}
\section{Conclusion and Future Works}
\label{sec:conclusion}
\vspace{-0.2cm}
We developed a model of human radiographic anatomy using simulated x-rays from a CT database and used the model to classify 20 radiographic landmarks. This lays the groundwork for C-arm positioning during time-sensitive procedures such as stroke thrombectomy. Translation into a clinical setting will require a 3D model that is robust to diverse patient populations and noisy imaging. Future work will expand the model to include rotational and depth information by annotating landmarks on the source CT scans and then training on a large sample of simulated x-rays taken at different C-arm angulation. The model will be further refined and evaluated on clinical fluoroscopy images acquired from biplane positioning during cerebral angiography. Finally, we will evaluate different trajectory strategies to minimize radiation dose and C-arm motion during procedures. 
%The proposed model was initially trained on a regression pretext task that develops an understanding of the skeletal anatomy. We train and test our model on data collected from NMDID, and it outperforms previous counterparts that are often used in similar tasks.}
\vspace{-8pt}
\section{Compliance with Ethical Standards}
\label{sec:ethics}
\vspace{-0.25cm}
This research study was conducted retrospectively using human subject data made available in open access by NMDID~\cite{edgar2020nmdid}. Ethical approval was not required as confirmed by the license attached with the open access data.
\vspace{-5pt}
\section{Acknowledgments}
\label{sec:acknowledgments}
\vspace{-0.25cm}
This work was supported by the National Science Foundation under Grant No. 2218063.
\vspace{-4pt}

% ------------------------------------------------------------------------- 
\bibliographystyle{ieeetr}
\scriptsize
\bibliography{refs}

\end{document}